%% file: main.tex
\documentclass{article}

\usepackage{microtype}
\usepackage{enumerate}
\usepackage{enumitem}
\usepackage{graphicx}
\usepackage{tabularx}
\usepackage{subfigure}
\usepackage{booktabs} % for professional tables
\usepackage{multirow} % For multi-row cells in tables
\usepackage{amssymb}
\usepackage{bbding}

\usepackage{hyperref}
\usepackage{booktabs} 
\usepackage{amssymb}    
\usepackage{pifont}

\usepackage[accepted]{icml2024}

\usepackage{amsmath}
\usepackage{amssymb}
\usepackage{mathtools}
\usepackage{amsthm}

\usepackage[capitalize,noabbrev]{cleveref}

\theoremstyle{plain}

\theoremstyle{definition}

\theoremstyle{remark}

\usepackage[textsize=tiny]{todonotes}
 
\newcommand{\ours}{POVID}

\icmltitlerunning{Aligning Modalities in Vision Large Language Models via Preference Fine-tuning}

\begin{document}

\twocolumn[
\icmltitle{Aligning Modalities in Vision Large Language Models via Preference Fine-tuning}

\icmlsetsymbol{equal}{*}
\begin{icmlauthorlist}
\icmlauthor{Yiyang Zhou}{equal,unc}
\icmlauthor{Chenhang Cui}{equal,unc}
\icmlauthor{Rafael Rafailov}{stf}
\icmlauthor{Chelsea Finn}{stf}
\icmlauthor{Huaxiu Yao}{unc}

\end{icmlauthorlist}

\icmlaffiliation{unc}{UNC-Chapel Hill}
\icmlaffiliation{stf}{Stanford University}

\icmlcorrespondingauthor{Huaxiu Yao}{huaxiu@cs.unc.edu}

\icmlkeywords{Machine Learning, ICML}

\vskip 0.3in
]
\printAffiliationsAndNotice{\icmlEqualContribution}

\input{abstract}

\input{introduction}

\input{preliminaries}

\input{method}

\input{experiment}

\input{related_work}

\input{conclusion}

\section*{Broader Impact Statement}
This paper presents work whose goal is to advance the field of machine learning and multimodal learning. There are many potential societal consequences of our work, none which we feel must be specifically highlighted here.

\section*{Acknowledgement}
We thank the Center for AI Safety for supporting our computing needs.
\bibliography{example_paper}
\bibliographystyle{icml2024}

%%%%%%%%%%%%%%%%%%%%%%%%%%%%%%%%%%%%%%%%%%%%%%%%%%%%%%%%%%%%%%%%%%%%%%%%%%%%%%%
%%%%%%%%%%%%%%%%%%%%%%%%%%%%%%%%%%%%%%%%%%%%%%%%%%%%%%%%%%%%%%%%%%%%%%%%%%%%%%%
% APPENDIX
%%%%%%%%%%%%%%%%%%%%%%%%%%%%%%%%%%%%%%%%%%%%%%%%%%%%%%%%%%%%%%%%%%%%%%%%%%%%%%%
%%%%%%%%%%%%%%%%%%%%%%%%%%%%%%%%%%%%%%%%%%%%%%%%%%%%%%%%%%%%%%%%%%%%%%%%%%%%%%%
\input{appendix}
\end{document}

%% file: abstract.tex
\begin{abstract}
Instruction-following Vision Large Language Models (VLLMs) have achieved significant progress recently on a variety of tasks. These approaches merge strong pre-trained vision models and large language models (LLMs). Since these components are trained separately, the learned representations need to be aligned with joint training on additional image-language pairs. This procedure is not perfect and can cause the model to hallucinate - provide answers that do not accurately reflect the image, even when the core LLM is highly factual and the vision backbone has sufficiently complete representations. In this work, we frame the hallucination problem as an alignment issue, tackle it with preference tuning. Specifically, we propose \ours\ to generate feedback data with AI models. We use ground-truth instructions as the preferred response and a two-stage approach to generate dispreferred data. First, we prompt GPT-4V to inject plausible hallucinations into the correct answer. Second, we distort the image to trigger the inherent hallucination behavior of the VLLM. This is an automated approach, which does not rely on human data generation or require a perfect expert, which makes it easily scalable. Finally, both of these generation strategies are integrated into an RLHF pipeline via Direct Preference Optimization. In experiments across broad benchmarks, we show that we can not only reduce hallucinations, but improve model performance across standard benchmarks, outperforming prior approaches. Our data and code are available at \url{https://github.com/YiyangZhou/POVID}.
\end{abstract}
\vspace{-0.5em}

%% file: introduction.tex
\vspace{-2em}
\section{Introduction}
\begin{figure}[t!]
  \centering
  \includegraphics[width=0.48\textwidth]{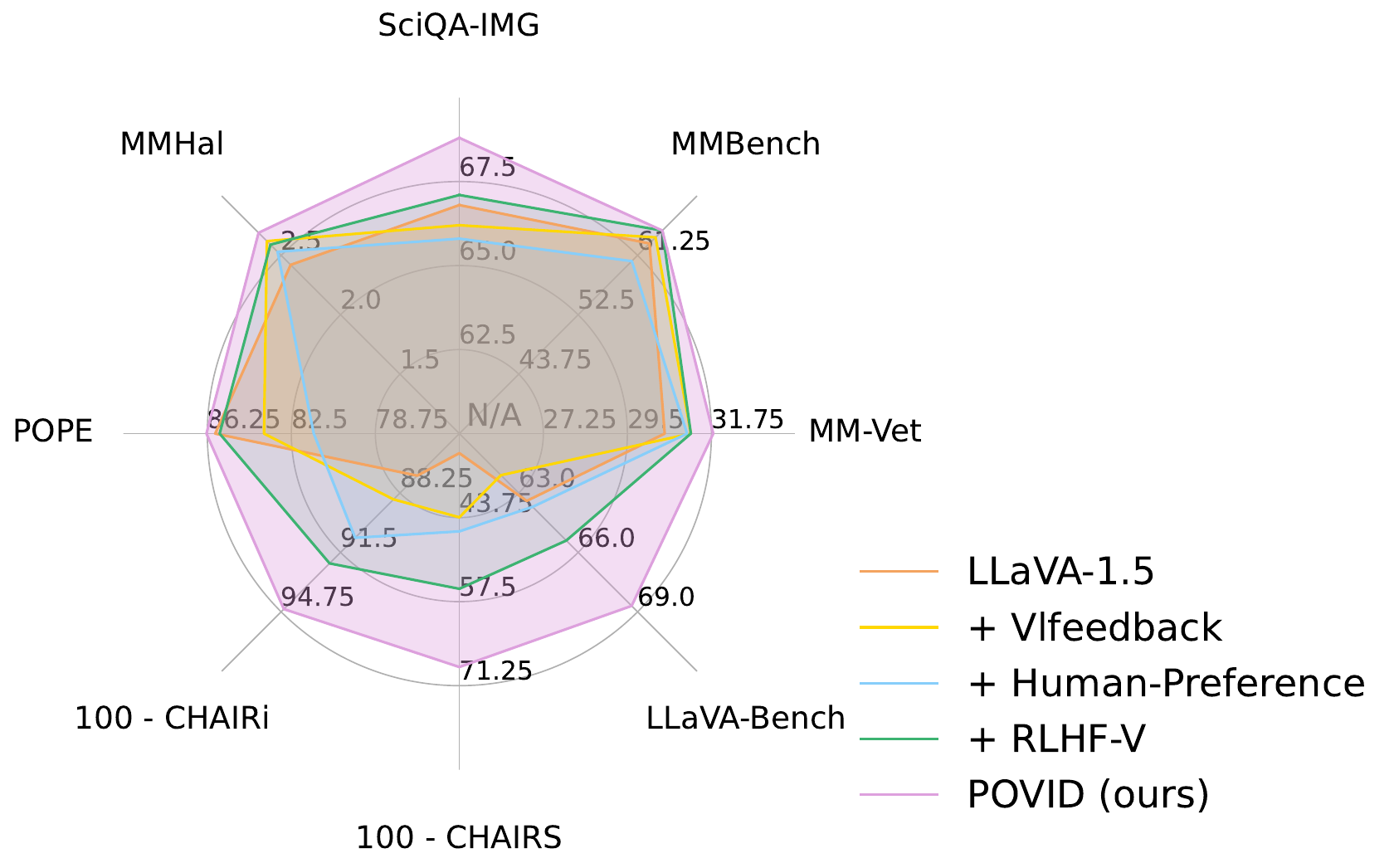}
    \vspace{-2em}
\caption{An overall performance comparison between \ours\ and other VLLM preference learning approaches.}
  \label{fig:radar_compare}
  \vspace{-1.5em}
\end{figure}

Vision Large Language Models (VLLMs) have achieved significant success in various vision understanding tasks, such as image captioning~\cite{vinyals2015show, li2022blip, li2023blip} and vision question answering~\cite{ye2023mplug, antol2015vqa}. These VLLM models fuse larger-scale pre-trained vision models into the representation space of a large language models (LLM), allowing the LLM access to the visual representations. However, such VLLMs are not perfect and even suffer from ``hallucinations", a phenomenon in which the language model generates content that is not grounded in the image, such as imagined objects and even scenes, wrong spatial relationships or categories, etc. Such artifacts are present even
when both the vision backbone produces high-quality visual
features and the language model itself is factual and accurate. These issues can pose significant risks when VLLMs are
deployed in high-stakes scenarios, such as medical domains~\cite{li2023llava} or autonomous driving~\cite{dewangan2023talk2bev}.

As discussed by~\citet{cui2023holistic}, the potential reason for hallucinations in VLLMs lies in their tendency to prioritize common sense or stereotypes present in the training language data, often disregarding the actual visual input information. In this paper, we attribute this issue to the lack of alignment between the image and text modalities, resulting in a reduced focus on input image information. Recent research efforts have sought to enhance the alignment between modalities through preference fine-tuning techniques, such as reinforcement learning from human feedback (RLHF)~\cite{sun2023aligning}. Concurrent works ~\cite{li2023silkie, zhao2023beyond} also use the Direct Preference Optimization (DPO) framework, but they rely on the traditional preference data generation process in LLMs, where both preferred and dispreferred responses may potentially be incorrect. However, in VLLMs, the produced responses are centered around the image data rather than being generated freely like in LLMs. When comparing two responses, both of which may be incorrect for the given task, the model may struggle to accurately align the image with the correct generated response. In \cite{2023rlhf-v} the authors propose to solve this issue by collection corrective feedback, which shows strong results, but relies on costly human data gathering.

Unlike prior works that generate both preferred and dispreferred data, we propose \textbf{P}reference \textbf{O}ptimization in \textbf{V}LLM with A\textbf{I}-Generated \textbf{D}ispreferences (\textbf{\ours}) framework, aiming to exclusively generate dispreferred feedback data using AI models. In \ours\, we employ a high-quality ground truth multi-modal instruction as the preferred answer and employ two strategies to generate dispreferred responses. \textit{First}, we utilize GPT-4V to introduce plausible hallucinations into the answer, which we then use as the dispreferred response. \textit{Second}, we aim to provoke inherent hallucination patterns and subsequently correct them within the target VLLM that requires fine-tuning. We achieve this goal by introducing noise, triggering inherent hallucination patterns within the VLLMs. The introduction of noise disrupts the VLLM's comprehension of the image, leading it to generate uncertain responses that rely more on textual context or the knowledge it has acquired from the training data. Given that the inherent hallucination patterns of the target VLLM evolve during the training process, the response generation with the noisy image occurs in real-time during training, and this is treated as dispreference. Finally, we integrate both forms of dispreference into the DPO optimization framework, specifically targeting the alignment of language generation with the image.

The primary contribution of this paper is \ours, which utilizes AI-generated dispreference to align the image and text modalities in VLLMs. This approach explicitly contrasts a hallucinatory answer with a truthful one, eliminating the need for gathering human feedback and making it easily deployable at scale. Our empirical results demonstrate the promise of our framework in reducing hallucinations and enhancing other VLLM-related tasks. In particular, as shown in Figure~\ref{fig:radar_compare}, our approach significantly improves performance compared to other preference tuning methods in VLLMs, achieving an average improvement of 12.4\% improvements on average. Additionally, we demonstrate that \ours\ can redirect the attention of VLLMs towards the image modality, resulting in better modality alignment.

\begin{figure*}[t!]
  \centering
  \includegraphics[width=0.99\textwidth]{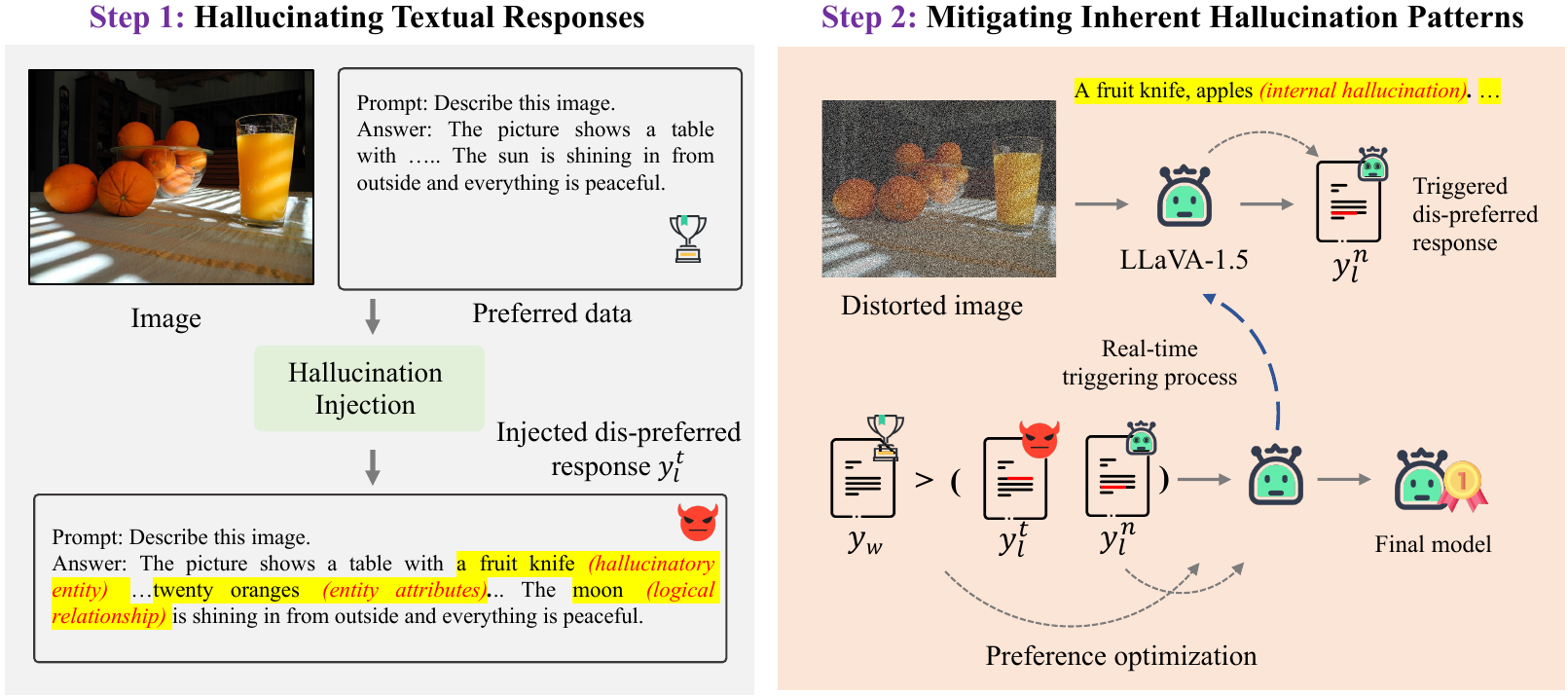}
\caption{The framework of \ours. The preference generation process is divided into two steps: hallucinating textual responses and trigger dispreference during training. Here, different types of triggered hallucinations are labeled in {\color{red}\textit{(types of hallucinations)}}.}
  \label{fig:framework}
\end{figure*}

%% file: preliminaries.tex
\section{Preliminaries}
Our approach aims to fine-tune VLLMs for better aligning the image and text modalities uses the framework of preference tuning from preferences over responses. In this section, we will provide some notations of VLLMs and an overview of direct preference optimization~\cite{rafailov2023direct}.

\noindent \textbf{Vision Large Language Models.} VLLMs is an multimodal extension of large language models, which can generate sentences in an autoregressive manner, aiming to progressively predict the probability distribution of the next token. Here, the input prompt $x$ contains both images and text prompts, and the output contains text response $y$. A typical application scenario for VLLMs is image captioning and Vision Question Answering (VQA).

\noindent \textbf{Direct Preference Optimization.} Reinforcement learning (RL) has shown its effectiveness in fine-tuning LLMs and align the LLMs behavior with human behavior. Typically, give an input $x$ and an output text response $y$, a language model policy $\pi_\theta$ can produce a conditional distribution $\pi_\theta(y\mid x)$. RL aims to maximize the average reward of outputs generated by the policy, where the reward function is defined as $r(x,y)$. To further avoid \textit{overoptimization}~\cite{gao2022scaling} happens when directly optimizing the reward function, the objective loss typically contains a KL-divergence term to control the divergence between the language model policy and its reference policy $\pi_{\text{ref}}(y\mid x)$ (e.g., the result after performing some supervised fine-tuning on LLMs). Thus, the overall objective is formulated as:
\begin{equation}
\label{eq:KLRL}
\max_{\pi_{\theta}}  \mathbb{E}_{x\sim \mathcal{D}, y\sim \pi_{\theta}(y \mid x)}\bigl[r(x, y) - \alpha\log \frac{\pi_\theta(y \mid x)}{\pi_\text{ref}(y \mid x)}\bigr]
\end{equation}
where $\mathcal{D}$ is a dataset of prompts and $\beta$ is a coefficient to control both terms. However, optimizing the above loss term with common strategies like proximal policy optimization (PPO)~\cite{schulman2017proximal} is complex to tune.

Recently, direct preference optimization (DPO)~\cite{rafailov2023direct} simplifies the above process by leveraging preference data for optimization. Here, the preference data is defined as $\mathcal{D}=\{x^{(i)}, y_w^{(i)}, y_l^{(i)}\}_{i=1}^N$, where $y_w^{(i)}$ and $y_l^{(i)}$ represent preferred and dispreferred responses given an input prompt $x$. Following a Bradley-Terry model~\citep{bradley1952rank}, the probably of obtaining each preference pair is:
\begin{equation}
    p(y_w\succ y_l)=\sigma(r(x, y_w)-r(x, y_l)),
\end{equation}
where we omit the superscript $(i)$ for simplicity and $\sigma(\cdot)$ is defined as a sigmoid function. In DPO, the optimization of Eqn.~\eqref{eq:KLRL} can be formulated as classification loss over the preference data as:
\begin{equation}
\begin{split}
&\mathcal{L}_{\textit{DPO}}(\pi_\theta; \pi_{\text{ref}}) = -\mathbb{E}_{(x,y_w,y_l) \sim \mathcal{D}} \\
&\left[ \log \sigma
\left(
\alpha \log \frac{\pi_\theta(y_w | x)}{\pi_{\text{ref}}(y_w | x)}
- \alpha \log \frac{\pi_\theta(y_l | x)}{\pi_{\text{ref}}(y_l | x)}
\right) \right].
\end{split}
\end{equation}
DPO enables learning $\pi_\theta$ from a fixed dataset of preferences, which is lightweight. However, the key challenge lies in generating effective preference data for fine-tuning and aligning image and text modalities in VLLMs.

%% file: method.tex
\begin{figure*}[t]
  \centering
  \includegraphics[width=0.91\textwidth]{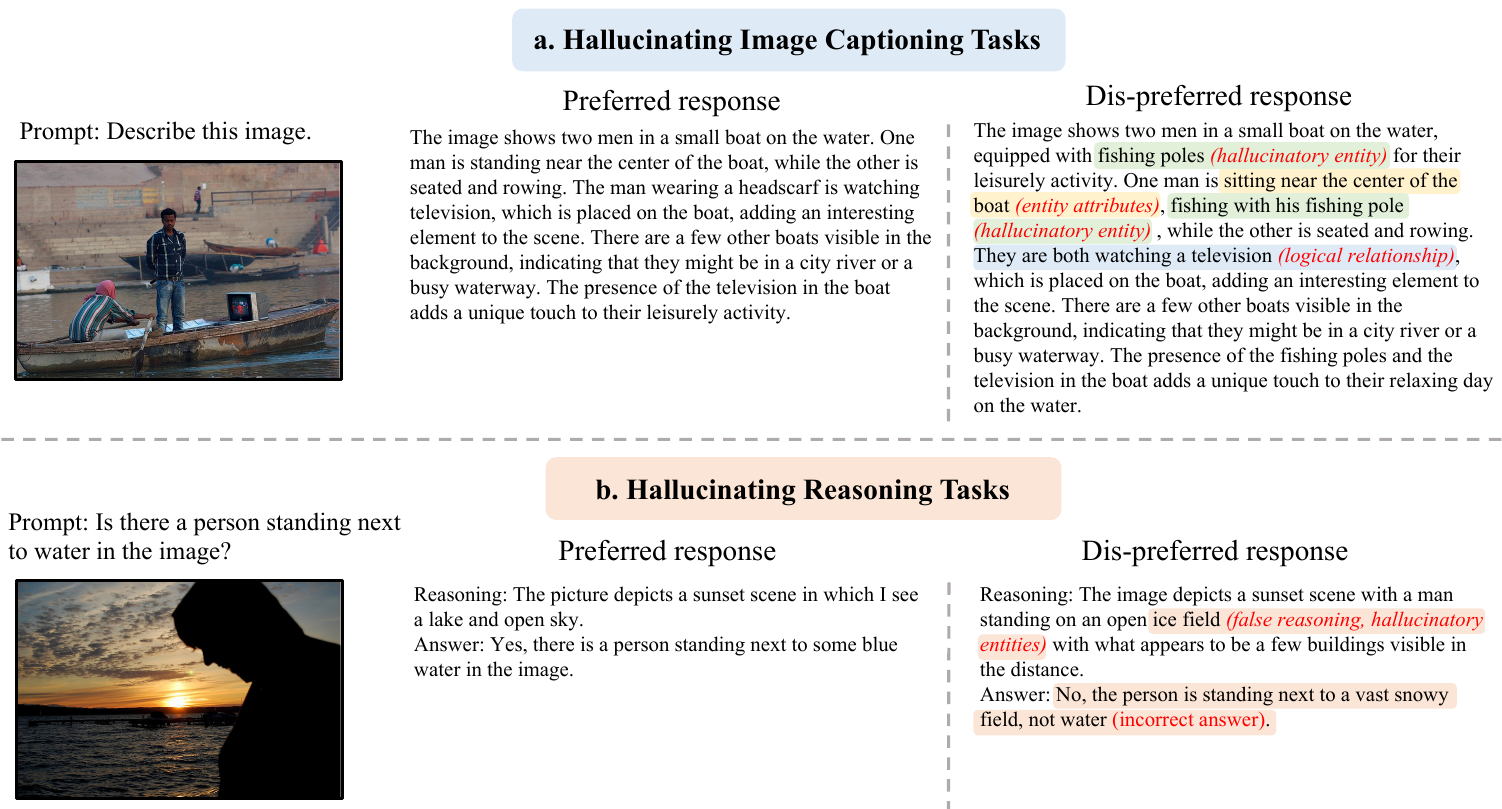}
\caption{Two examples extracted from hallucinated image captioning tasks and reasoning tasks. Different types of hallucinations are labeled in {\color{red}\textit{(types of hallucinations)}}.}
  \label{fig:data1}
  \vspace{-1.5em}
\end{figure*}

\section{Constructing Preferences to Aligning Modalities in VLLMs}
While preference learning approaches (e.g., DPO) facilitate the lightweight and stable training of VLLMs, they require data in the form of preferences. In contrast to LLMs, which support more freestyle generation in many scenarios, VLLMs used in various applications, such as VQA or image captioning, produce responses linked to input images. This inherent image-centricity presents distinct challenges in the preference data generation process for VLLMs, setting it apart from the process in LLMs. Specifically, in VLLMs, when comparing two responses, neither of which is correct for the required task (e.g., image captioning), the model may not be able to accurately align the image with the response.

To address this challenge, we propose \textbf{P}reference \textbf{O}ptimization in \textbf{V}LLM with A\textbf{I}-Generated \textbf{D}ispreferences (\textbf{\ours}), a novel approach aimed at better aligning image and text modalities. As illustrated in Figure~\ref{fig:framework}, \ours\ leverages AI models to generate dispreferred responses without the need for human labeling efforts. These generated dispreferred responses, when combined with groundtruth image descriptions (treated as preferred responses), form the preference data pairs. Specifically, we employ two strategies to generate the dispreferred response: (1) Firstly, we manipulate the groundtruth text response by transforming the groundtruth response into hallucinated response, which serves as the dispreferred response; (2) Secondly, we introduce distortion to the image input during the training process, intending to trigger inherent hallucination patterns within the VLLMs. These patterns are then formalized as the dispreferred response, motivating the model to correct its inherent dispreferred patterns. In the remainder of this section, we will provide detailed explanations of both strategies and demonstrate how to integrate them into the preference training framework.

\subsection{Hallucinating Textual Responses}
\label{sec:data_collection}

\begin{figure}[t]
  \centering
  \includegraphics[width=0.45\textwidth]{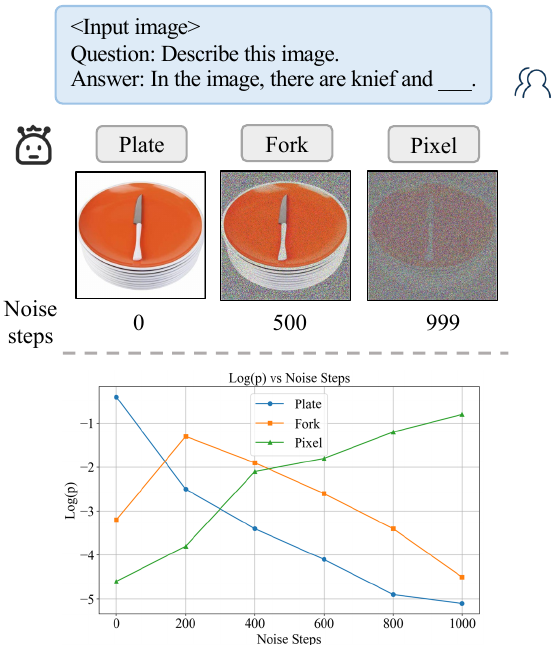}
\caption{Illustration of logits for the next token generation with ``In the image, there are knife and \_". This figure shows the predictive uncertainty in token generation, emphasizing the influence of visual cues from objects identified as ``knife" and ``plate".}
  \label{fig:diffusionnoise}
  \vspace{-1.5em}
\end{figure}

In our first strategy, we aim to generate dispreferred hallucinatory responses by hallucinating the groundtruth correct response. We construct the hallucinatory response based on a subset with 17K examples that are randomly sampled from LLaVA-Instruct-150K~\cite{liu2023visual} dataset.  Here, the LLaVA-Instruct-150K datasets is used to train LLaVA LLaVA with supervised fine-tuning. The 17K examples includes various task types, including image captioning, simple VQA and logical reasoning. 

To construct the preferences, we treat the original answers in the 17K examples as preferred responses. In terms of constructing dispreferred responses, we hallucinate the original answers using GPT-4V~\cite{OpenAI_GPT4_2023}. Here, we adopt two hallucinating approaches tailored to different tasks:

\textbf{I. Hallucinating Image Captioning Tasks.} First, we hallucinate the image captioning tasks by considering three fundamental causes of hallucination in VLLMs: (1) \emph{Object Co-occurrence}: This phenomenon arises when the training data contains spurious co-occurring patterns between objects, leading VLLMs to generate objects based on these learned spurious correlations. In this context, we aim to leverage GPT-4V to deduce object co-occurrence within the given image and subsequently revise the original responses accordingly; (2) \emph{Logical Relationships Between Entities}: This involves using GPT-4V to modify the relationships between the original objects; (3) \emph{Incorrect Attributes}: In this case, we employ GPT-4V to alter the attributes of various objects, such as changing their colors. We illustrate these three distinct hallucination scenarios with an example provided in Figure~\ref{fig:data1}(a). In addition, the prompt we used to generate the dispreferred response is in Appendix~\ref{sec:prompt}.

\textbf{II. Hallucinating Reasoning Tasks.} Secondly, when dealing with tasks involving reasoning, such as VQA and logical reasoning, we task GPT-4V with modifying the reasoning process. This entails introducing errors related to logical relationships, entity information, entity attributes, and more. Additionally, we recommend that GPT-4V attempts to make subtle changes to the reasoning process, ensuring it remains independent of factual reasoning results, meaning that an incorrect reasoning process may still yield correct results. However, if the introduction of errors necessitates alterations to the reasoning results, we instruct GPT-4V to adjust the results accordingly. Likewise, in Figure~\ref{fig:data1}(b), we provide an example to demonstrate both the original and the generated dispreferred responses. The prompt we used is detailed in Appendix~\ref{sec:prompt}.

\subsection{Mitigating Inherent Hallucination Patterns}
\label{sec:image_distortion}
In addition to generating the dispreferred response using powerful external AI models like GPT-4V, we also aim to provoke inherent hallucination patterns within the VLLM to be finetuned. Our second strategy introduces noise into the image to trigger inherent hallucination patterns in the VLLMs. This noise disrupts the VLLM's understanding of the image, leading it to produce uncertain responses that rely more on textual context or acquired knowledge from the training data. This occurs because, in the presence of noisy images, the model tends to prioritize inherent object associations over visual information. Notably, the noise step should remain within a reasonable range, ensuring that the image remains easily recognizable by humans. For example, as depicted in Figure~\ref{fig:diffusionnoise}, when presented with the context ``There are a knife and \_", under specific noisy conditions, the likelihood of ``fork" surpasses that of ``plate" (ground truth). This may occur because ``plate" is more likely to co-occur with ``fork" in the training data. With an increase in noise steps, the term ``pixel" becomes predominant, owing to the noticeable noise patterns within the image. Consequently, establishing an appropriate noise step to activate inherent hallucination patterns is a reasonable approach.

To achieve this goal, we introduce diffusion noise into the original image. We define the noise step as $k$, and the noised image with step $k$ can be expressed as follows:
\begin{equation}
    x(k)=\sqrt{\bar{\xi}_k} \cdot x + \sqrt{1 - \bar{\xi}_k} \cdot \epsilon, 
\end{equation} 
where $\bar{\xi}_t = \prod_{i=0}^{k} \xi_i$ and $ \xi_k \in (0, 1)$ is a hyperparameter chosen prior to model training. Detailed settings can be found in Appendix~\ref{sec:setup}. After obtaining the noised image, in order to more effectively capture changes in inherent hallucination patterns during the fine-tuning process of the VLLM, we integrate the image noising process into the DPO fine-tuning process. Specifically, for each input prompt $x$, we take into account the dispreferred responses from both the hallucinated text responses discussed in Section~\ref{sec:data_collection} and the responses triggered by distorted images. We then reformulate the DPO loss as follows:

\begin{equation}
\small
\label{eq:self_loss}
\begin{split}
&\mathcal{L}_{\textit{\ours}}(\pi_\theta; \pi_{\text{ref}}) = -\mathbb{E}_{(x,y_w,y_l) \sim \mathcal{D}}
\Bigg[ \log \sigma \Bigg(  \alpha \log \frac{\pi_\theta(y_w | x)}{\pi_{\text{ref}}(y_w | x)}  
\ 
 \\
&-  \left( \beta_1 \log \frac{\pi_\theta(y^t_l | x)}{\pi_{\text{ref}}(y^t_l | x)} + \beta_2 \log \frac{\pi_\theta(y^{n}_l | x^{n})}{\pi_{\text{ref}}(y^n_l | x^{n})} \right) \Bigg) \Bigg] ,
\end{split}
% \label{dpoself}
\end{equation}

where $\alpha$, $\beta_1$ and $\beta_2$ are coefficients that balance preferred and dispreferred terms. $y_l^g$ represents the dispreferred response generated using the approach outlined in Section~\ref{sec:data_collection}. Additionally, $x^{n}$ represents the noisy image, which triggers the generation of the dispreferred response $y_l^{n}$. It's important to note that for each token $i$ in the sequence $y_l^{n}$, the value of $y^{n}_{l,i}$ is determined by selecting the maximum probability from the set ${\pi_{\theta}{(\cdot \mid x^{n}, y_{w,<i})}}$. Here, each generated token in the dispreferred response $y^{n}_{l,i}$ is conditioned on the prior tokens from the preferred response $y_{w,<i}$. This conditioning allows us to control the reliability of the triggered dispreferred response. As a result, we aim to capture the most significant changes between the preferred and dispreferred responses, since a substantial portion of dispreferred response overlaps with preferred response. The training process of our method is detailed in Algorithm \ref{al_ps}.

\begin{algorithm}
\caption{\ours\ Training Process}
\begin{algorithmic}[1]
\REQUIRE
 % \textbf{Input:}
 $\mathcal{D}$: Dataset of paired images and text context.
 $\pi_\theta$: Parameters of the VLLM.
$\pi_{\text{ref}}$: Parameters of the reference model.
 $\alpha$, $\beta_1$, $\beta_2$: Hyperparameters.
 $\xi_k$: Noise hyperparameter for each timestep. $T$: Noise Steps
\STATE{AddNoiseToImage}{($x_0, k$)} \\ 
$\epsilon \sim \text{N}(0,1)$  \\
 $x(k) \gets \sqrt{\bar{\xi}_k} \cdot x_0 + \sqrt{1 - \bar{\xi}_k} \cdot \epsilon$

\STATE Generate disprefered data and place it in $\mathcal{D}$

\STATE Initialize reference policy $\pi_\theta$
\FOR{epochs}
\FOR{$(x, y_w, y^t_l) \in \mathcal{D}$}
\FOR{$k = 0$ to $T$} 
\STATE $x({k}) \gets \text{AddNoiseToImage}(x, k)$

\ENDFOR

\STATE Update $\pi_\theta$ through Eqn.~\eqref{eq:self_loss} 
\ENDFOR
\ENDFOR

\end{algorithmic}
\label{al_ps}
\end{algorithm}

%% file: experiment.tex
\begin{table*}[!t]
\small
\centering
\renewcommand\arraystretch{1.2}
\caption{Comparison between \ours\ and other preferences construction approaches in both hallucination and comprehensive evaluation benchmarks. We bold the best results and underline the second best results.}
\vspace{0.4em}

\resizebox{\textwidth}{!}{\setlength{\tabcolsep}{1.5mm}{

\begin{tabular}{l|cccc|cccc}
\toprule
 & \multicolumn{4}{c|}{Hallucination Benchmark} & \multicolumn{4}{c}{Comprehensive Benchmark} \\ \midrule
  Method & CHAIR$_{S}$ $\downarrow$  & CHAIR$_{i}$ $\downarrow$ & POPE $\uparrow$ & MMHal $\uparrow$ & SciQA-IMG $\uparrow$ & MM-Vet $\uparrow$ & MMBench $\uparrow$ & LLaVA-Bench $\uparrow$   \\\midrule

LLaVA-1.5  & 66.8     & 12.7 & 85.90 & 2.42 & 66.8 & 30.5 & 63.0 & 63.4        \\  
+ Vlfeedback & 56.3 & 11.4  & 83.72 & \underline{2.62} & 66.2 & \underline{31.2} & \underline{63.9} & 62.1 \\
+ Human-Preference & 54.0 & 9.3 & 81.50 & 2.53 & 65.8    & 31.1 & 60.4 &  63.7\\
+ RLHF-V & \underline{44.6} & \underline{7.9} &  \underline{86.20}& 2.59 & \underline{67.1}   & 30.9& 63.6 & \underline{65.4} \\\midrule

\textbf{\ours\ (ours)}   & \textbf{31.8}      & \textbf{5.4} &  \textbf{86.90} & \textbf{2.69} & \textbf{68.8}  & \textbf{31.8} & \textbf{64.9} & \textbf{68.7}  \\
\bottomrule
\end{tabular}}}
\vspace{-1em}
\label{chair}
\end{table*}

\begin{table*}[h]
\small
\centering
\renewcommand\arraystretch{1.2}
\caption{Comparison between \ours\ and other state-of-the-art VLLMs across both hallucination and comprehensive evaluation benchmarks. We bold the best results and underline the second best results.}
\vspace{0.4em}
\resizebox{\textwidth}{!}{\setlength{\tabcolsep}{1.5mm}{

\begin{tabular}{l|cc|ccccc}
\toprule

Method& Vision Encoder & Language Model  & CHAIR$_{S}$ $\downarrow$  & CHAIR$_{i}$ $\downarrow$  & 
POPE $\uparrow$ & MMHal $\uparrow$ & Avg. ranking $\downarrow$  \\  \midrule

InstructBLIP     & ViT-g (1.3B) & Vicuna (7B)    & \underline{40.0}     & \underline{8.0} & 77.83 & 2.10 & 3.00\\
Qwen-VL-Chat     & ViT-G (1.9B) & Qwen (7B)    & 48.2     & 9.1 & \textbf{87.07} & \textbf{2.89} & \underline{2.50}\\
mPLUG-Owl2     & ViT-L (0.3B) & LLaMA (7B)    & 54.4     & 12.0 & 86.20 & 2.17& 4.00 \\\midrule
  \textbf{\ours\ (ours)}     & ViT-L (0.3B) & Vicuna (7B)    & \textbf{31.8}     & \textbf{5.4} & \underline{86.29} & \underline{2.69}   & \textbf{1.50}\\  
\midrule
\midrule
 
 Method& Vision Encoder & Language Model & SciQA-IMG $\uparrow$ & MM-Vet $\uparrow$ & MMBench $\uparrow$ & LLaVA-Bench $\uparrow$ & Avg. ranking $\downarrow$ \\ 
\midrule 
InstructBLIP & ViT-g (1.3B) & Vicuna (7B) & 60.5  & 26.2 & 36.0 & 60.9 & 4.00\\
Qwen-VL-Chat & ViT-G (1.9B) & Qwen (7B) & \underline{68.2} & \textbf{41.2} & 60.6 & \underline{67.7} & \underline{2.25}\\ 
mPLUG-Owl2 & ViT-L (0.3B) & LLaMA (7B)  & 64.5 & \underline{36.2} & \underline{64.5} & 59.9 & 3.00\\\midrule
\textbf{\ours\ (ours)} & ViT-L (0.3B) & Vicuna (7B) & \textbf{68.8}  & 31.8 & \textbf{64.9} & \textbf{68.7} & \textbf{1.75}\\
%LLaVA-1.5\textsubscript{DPO\textsubscript{self}}
\bottomrule
\end{tabular}}}
\vspace{-1em}
\label{comprehensive}
\end{table*}

\section{Experiment}
In this section, we empirically investigate the effectiveness of \ours\ in aligning image and text modalities in VLLMs and reducing hallucination. We aim to answer the following questions: (1) Can \ours\ effectively reduce hallucination in VLLMs compared to other preference fine-tuning strategies? (2) Does \ours\ improve performance compared to other benchmarks and tasks like VQA? (3) Can hallucinating textual responses and image distortion benefit performance? (4) How does \ours\ change attention weights to align image and text modalities?

\subsection{Experimental Setups}
In this section, we briefly introduce the implementation details, baselines, and evaluation settings.

\noindent \textbf{Implementation Details.}
Following concurrent VLLM preference tuning studies~\cite{yu2023rlhf, li2023silkie}, we have chosen LLaVA-1.5 (7B) as our backbone model for all experiments and have applied \ours\ to fine-tune LLaVA-1.5 (7B). The overall training process is divided into two stages. In the first stage, we exclusively utilize the preferences generated through the hallucinating textual responses, as discussed in Section~\ref{sec:data_collection}, to fine-tune LLaVA-1.5 using DPO. In the second stage, we employ image distortion to rectify the model's inherent hallucinatory behaviors using the loss defined in Eqn.~\eqref{eq:self_loss}. The first stage involves training for 3 epochs, and the second stage for 1 epoch. The entire training process requires a single A100 80GB GPU and takes approximately 6 hours. For a more detailed description, please refer to Appendix~\ref{sec:setup}.

\textbf{Baseline Approaches.}
We first compare the proposed approach with other VLLM preference tuning methods, which include Silkie~\cite{li2023silkie}, LLaVA-RLHF~\cite{sun2023aligning}, and RLHF-V~\cite{yu2023rlhf}. These methods share a common goal of enhancing model performance by creating curated datasets and subsequently applying preference tuning techniques to fine-tune the model based on these datasets. To ensure a fair and equitable comparison, we utilize the same curated datasets employed by these approaches and apply DPO to fine-tune LLaVA-1.5 (7B).

Furthermore, we compare the performance with other open source VLLMs, including InstructBLIP \cite{instructblip}, Qwen-VL-Chat \cite{bai2023qwen} and mPLUG-Owl2 \cite{ye2023mplug}. 

\begin{table*}[!t]
\small
\caption{Results of ablation study. Text disprefer means we only using hallucinated textual response to train DPO. Image distortion means that we use distorted images to trigger inherent hallucination patterns.}
\centering
\vspace{0.5em}
\resizebox{\textwidth}{!}{\setlength{\tabcolsep}{1.5mm}{
\begin{tabular}{@{}cc|cccc|cccc@{}}
\toprule
&& \multicolumn{4}{c}{Hallucination Benchmarks}& \multicolumn{4}{|c}{Comprehensive Benchmarks}\\ \midrule
Text disprefer & Image distortion & CHAIR$_{S}$ $\downarrow$ &  CHAIR$_{i}$ $\downarrow$ &POPE $\uparrow$ & MMHal $\uparrow$ & SciQA-IMG $\uparrow$ & MM-Vet $\uparrow$ & MMBench $\uparrow$ & LLaVA-Bench $\uparrow$ \\
\midrule
  \texttimes & \texttimes & 66.8 & 12.7 & 85.90&2.42 & 66.2 & \underline{31.2} & 63.9 & 62.1 \\
  \checkmark & \texttimes & \underline{39.6} & \underline{6.3} & \underline{86.04}&\underline{2.65} & \underline{67.2}&30.9&\underline{64.7}&\underline{67.5} \\
  \texttimes & \checkmark & 50.4 & 9.6 &85.19& 2.54 & 66.9&30.7&64.3&66.9 \\
 \checkmark & \checkmark & \textbf{31.8} & \textbf{5.4} &\textbf{86.90} &\textbf{2.69} & \textbf{68.8}&\textbf{31.8}&\textbf{64.9}&\textbf{68.7} \\
\bottomrule
\end{tabular}}}
\label{table:ablation_study} 
\vspace{-1.5em}
\end{table*}

\textbf{Evaluation Benchmark.}
To evaluate the performance of \ours\ and other baselines, we first adopt VLLM hallucination evaluation benchmarks, including CHAIR~\cite{rohrbach2018object}, POPE~\cite{li2023evaluating}, and MMHal~\cite{sun2023aligning}. Here, CHAIR, including CHAIR$_{S}$ and CHAIR$_{I}$, is a metric used in image captioning tasks to evaluate the accuracy of object descriptions in captions. It compares the objects mentioned in a caption with those present in the image. 
MMHal~\cite{sun2023aligning} assesses hallucinations and response informativeness by utilizing GPT-4V to compare model output with human responses and various object labels, determining the scores accordingly.
POPE~\cite{li2023evaluating} uses a set of binary classification tasks, prompting VLLMs with simple Yes-or-No questions about the existence of certain objects in images.

We further evaluate all approaches on comprehensive VLLM evaluation benchmarks, including SciQA-IMG~\cite{lu2022learn}, MME~\cite{fu2023mme}, MMbench \cite{liu2023mmbench}, MM-Vet \cite{yu2023mm} and LLaVA-bench\cite{liu2023visual}. Each benchmark contains tasks to evaluate perception, cognition, and reasoning abilities of VLLMs. More detailed descriptions of these baselines are in Appendix~\ref{sec:base}.

\subsection{Results}
\textbf{Comparison with Different Preferences in VLLMs.}
In Table \ref{chair}, we present the results of a comparison between various VLLM preferences, evaluating both hallucination and comprehensive benchmarks. Firstly, in the hallucination benchmarks, \ours\ effectively enhances performance by creating dispreferred preferences through textual data manipulation and image distortion. We achieve a significant improvement of 31.78\% across all hallucination benchmarks, effectively reducing hallucinations in the generated responses. This outcome aligns with our expectations, as constructing dispreferences from the ground-truth correct responses maximally enables the model to discern differences between correct and incorrect responses while optimizing alignment between the image and text modalities within the model. Moreover, in more comprehensive evaluation benchmarks, which encompass not only factuality and hallucination assessment but also other aspects, \ours\ continues to demonstrate superior performance when compared to other preference data collection methods. This further indicates our model's capacity to enhance VLLM performance through improved modality alignment.

\textbf{Comparison with Open-Sourced VLLMs Models.}
We present a comparison between \ours\ and other open-sourced VLLMs in Table~\ref{comprehensive}. Although various approaches utilize different image and text encoders, \ours\ outperforms other popular VLLMs in five out of eight benchmarks. In contrast, the second-best baseline, Qwen-VL-Chat, achieves the best performance in only two out of eight benchmarks. This underscores the superiority of \ours\ and further corroborates its effectiveness in aligning image and text modalities to improve the performance of VLLMs.

\begin{figure*}[t!]
  \centering
  \includegraphics[width=0.95\textwidth]{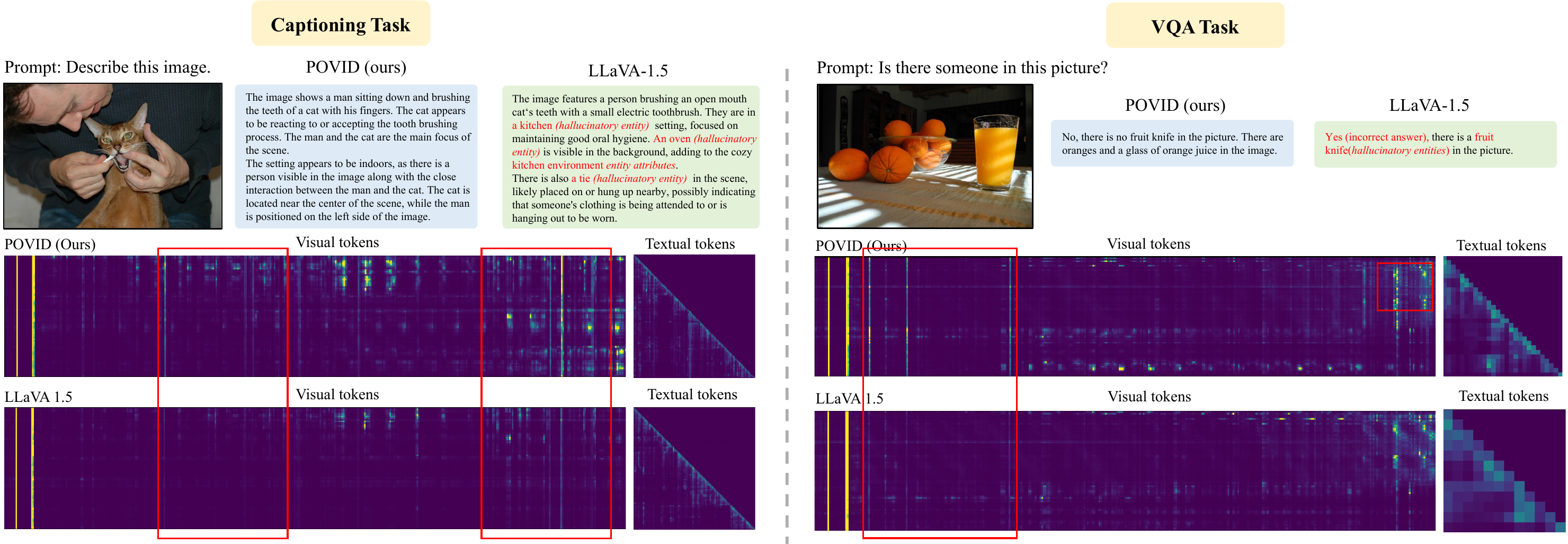}
\caption{Comparison of attention map between \ours\ and LLaVA-1.5 at different tasks. The red box region is labeled with the image attentions that can be significantly improved by \ours.}
\vspace{-0.5em}
  \label{fig:att1}
  \vspace{-1.2em}
\end{figure*}

\subsection{Analysis}
In this section, we provide a comprehensive analysis to demonstrate how different components contribute to the performance of \ours\ and illustrate how \ours\ enhances overall performance.

\textbf{Ablation Studies.}
To further demonstrate the essential role of the key components of \ours\ in contributing to performance improvement, we conducted ablation experiments on both hallucination and comprehensive benchmarks, and present the results in Table~\ref{table:ablation_study}. In this ablation study, we evaluate the effectiveness of two aspects: (1) hallucinating groundtruth responses and (2) image distortion. According to the results, we initially observe that image distortion can enhance performance across all benchmarks. This indicates its effectiveness in aligning multimodalities by compelling the model to rectify inherent hallucination patterns. Additionally, generating dispreference from groundtruth responses significantly enhances performance, underscoring the effectiveness of the AI-generated dispreference strategy. Finally, when combining both strategies, \ours\ achieves the best performance, further affirming its effectiveness in enhancing VLLMs through improved modality alignment.

\textbf{Fine-grained Performance Analysis.}
Table~\ref{table:models_comparison_finegrained} presents a fine-grained performance analysis of different preference collection strategies on the LLaVA-Bench benchmark. This analysis encompasses a spectrum of multi-modal reasoning and perception dimensions, such as Conversation, Detail Description, and Complex Reasoning. According to Table~\ref{table:models_comparison_finegrained}, it is evident that, when compared with other preference data collection approaches, \ours\ excels in image captioning and providing detailed descriptions for a given image. This outcome aligns with our expectations, as our training data includes various long-form captions, and such comprehensive preference comparisons result in improved alignment and stronger image captioning results. Moreover, across other categories, our approach consistently outperforms alternative preference collection strategies, underscoring its effectiveness across various tasks.
\vspace{-1em}

\begin{table}[h]
\centering
\caption{Fine-grained performance comparison of various models on LLaVA-Bench, where we adopt the following abbreviation: Convo for Conversation, Captioning for Detail description, Reasoning for Complex reasoning.}
\vspace{0.5em}
\resizebox{\columnwidth}{!}{\setlength{\tabcolsep}{1.5mm}{
\begin{tabular}{l|cccc}
\toprule
Method & Convo & Captioning & Reasoning & Overall \\
\midrule
LLaVA-1.5 & 53.3 & 53.4 & 79.6 & 63.4 \\
+ Vlfeedback & 51.3 & 49.3 & 78.5 & 62.1 \\
+ Human-Preference & 49.6 & 43.3 & \underline{81.3} & 63.7 \\
+ RLHF-V & \underline{55.8} & \underline{56.1} & 80.3 & \underline{65.4} \\\midrule
\textbf{\ours\ (ours)} & \textbf{55.9} & \textbf{60.1} & \textbf{81.5} & \textbf{68.7} \\
\bottomrule
\end{tabular}}}
\vspace{-1em}
\label{table:models_comparison_finegrained}
\end{table}

\textbf{Modality Alignment Analysis.}
We assess the impact of \ours\ on modality alignment by comparing the attention maps generated by \ours\ with those of the original LLaVA-1.5 model, with a specific focus on image captioning and VQA tasks. We illustrate two cases in Figure~\ref{fig:att1}, where these attention maps reveal the distribution of attention scores assigned to generated textual tokens within the input image-text sequence throughout the VLLM's output generation phase. Our findings reveal that the original LLaVA-1.5 model tends to overemphasize the context of the text, which can result in hallucinations. In contrast, \ours\ increasingly prioritizes attention towards the image, indicating a strong alignment between image and text modalities. One potential explanation for this phenomenon is that, through a comparison between the ground truth and the generated dispreferred data, along with the mitigation of internal hallucination patterns, \ours\ redirects the VLLM's attention, leading to a greater focus on the image tokens.

%% file: related_work.tex
\vspace{-0.5em}
\section{Related Work}

\textbf{VLLMs and VLLM Hallucination.}
The advent of autoregressive large-scale language models (LLMs), highlighted in works by \cite{touvron2023llama, touvron2023llama2, alpaca}, has led to the development of Vision-Large Language Models (VLLMs). 
To align the image and text modalities, recent research has concentrated on instruction tuning~\cite{li2023visionlanguage}, scaling up training dataset~\cite{jia2021scaling}, and better alignment between image and text with local feature enhancement~\cite{cha2023honeybee}. 
These advancements have successfully combined LLMs with image inputs and excel in image comprehension. 
% However, VLLMs can sometimes experience  hallucinations,  generating outputs that may not accurately or faithfully represent the content of a user-provided image. 
However, such VLLMs are
 not perfect and even suffer from “hallucinations”, generating outputs that may not accurately or faithfully represent the content of a user-provided image. There are various sources of hallucinations
in VLLMs, including biased data \cite{chuang2023debiasing, tu2023many}, insufficient training \cite{chen2023lion}, and imperfect 
 inference \cite{huang2023opera}.
Recently, addressing  hallucination in LVLMs is
primarily achieved through various techniques such as  decoding 
approaches \cite{leng2023mitigating, huang2023opera}, post-processing \cite{zhou2023analyzing, yin2023woodpecker} and the construction of higher-quality dataset \cite{liu2023aligning, li2023m}. While these approaches can mitigate hallucination to some extent, they often fail to directly guide VLLMs to align image and text modalities.

\textbf{Preference Alignment}
Aligning with human preferences for large models has emerged as a critical issue due to the limitations imposed by safety and ethical considerations in real-world applications.
Preference alignment can be broadly categorized into two main approaches: alignment through feedback, which encompasses both human \cite{bai2022training, rafailov2023direct} and AI-generated feedback~\cite{lee2023rlaif} and alignment via prompt guidance~\cite{wei2022chain}. Initial investigations into preference alignment for VLLMs have recently been conducted. \citet{sun2023aligning} introduced LLaVA-RLHF, which utilizes a preference dataset annotated by humans to decrease hallucinations in LLaVA.  \citet{li2023silkie} proposed a method for distilling preferences into VLLMs to enhance their ability to generate relevant and accurate responses based on visual context. \citet{yu2023rlhf} collected human preferences in the form of segment-level corrections to hallucinatory content and optimizing the model's behavior based on dense, direct feedback. While these initial results are promising, these works heavily rely on the traditional preference data generation process in LLMs, which generate both preferred and dispreferred responses, but none of them are guaranteed to be correct. In VLLMs, when both responses prove incorrect for the given task, accurately aligning the image with the correct generated response becomes challenging. In contrast, \ours\ directly generates dispreferred responses, effectively addressing this challenge.

%% file: conclusion.tex
\vspace{-0.5em}
\section{Conclusion}
In this work, we introduce a novel approach, Preference Optimization in VLLM with AI-Generated Dispreferences (\ours) to address the challenges in modality alignment for large vision-language models. In \ours, we adopt two strategies to generate disprefered responses: first, we use synthetic data from GPT-4V to inject plausible hallucinations into the correct answer. Second, we use distorted images to trigger the inherent hallucination behavior of the VLLM. Then both of these answers are integrated into an RLHF framework via Direct Preference Optimization. Empirical evaluations across multiple benchmarks reveal that \ours\ not only mitigates hallucination effectively but boosts the overall performance of model.

%% file: appendix.tex
\newpage
\appendix
\onecolumn

\section{Appendix.}

\subsection{Training Setup}
\label{sec:setup}
Training hyperparameters are shown in Table \ref{hp}. For the first phase, we trained for 3 epochs, and for the second phase, the training was conducted for 1 epoch. Training for 20 hours on one A100 80G GPU.
For the second phase, we adjust the diffusion noise level, symbolized by 
$\xi$ through a specific formula: $\xi = \text{Sigmoid}(l_t) \times (0.5 \times 10^{-2} - 10^{-5}) + 10^{-5}$, where $\epsilon$ is drawn from a normal distribution.

\subsection{Construction of the Dispreference Dataset}
\label{sec:prompt}
This section details the prompts utilized to compile the dataset focusing on dispreferences, specifically within the realms of image captioning and reasoning tasks. The prompts are designed to elicit responses that reveal dispreference patterns, categorized into two main types: image captioning tasks intended to provoke imaginative descriptions, and reasoning tasks aimed at stimulating inferential thought processes. These prompts, central to our methodology, are enumerated in Table \ref{prompt}, offering a comprehensive view of the data generation framework.

\begin{table}[h]
\centering
\caption{Training hyperparameters.}
\begin{tabular}{lc}
\toprule
Hyperparameters & \\ \midrule
lora\_r                   &       128 \\
lora\_alpha                      &       256 \\
lora\_target & all\\
mm\_projector\_lr         &       2e-5 \\
Batch size  &   1 \\
Learning rate & 1e-7 \\
model\_max\_length & 1024\\
noise\_step (only for internal preference optimization) & 500\\ 
\bottomrule
\end{tabular}
\label{hp}
\end{table}

\begin{table}[h]
    \centering
        \caption{Two types of prompts to GPT4V (The format of the obtained data is \{image, prefer data, disprefer data\}).}
        \vspace{0.5em}
    \setlength{\tabcolsep}{10pt}
    \renewcommand{\arraystretch}{1.2}
    \begin{tabularx}{\textwidth}{X}
        \toprule
        \textbf{Prompts for hallucinating image captioning tasks:} \\Help me generate one highly confusing response based on the image and the standard caption in the Question-Answer Pair.\\
        *****************************************\\
            Question-answer Pair:\\
            Q: \{question\}\\
            A: \{answer\}\\
            Requirements:\\
            (1) The generated caption is generally similar to the given A, with the same main meaning; (2) You can refer to the following errors to generate the wrong caption (1. The wrong caption can contain some co-occurring objects, which are prone to appear in such scenarios but do not appear in the image; 2. The wrong caption can be an error in the number of entities or the logical relationships between entities; 3. The attributes of entities in the caption can also be modified, such as color, appearance, etc.) (3) Compared to the original caption A, the caption you modified is incorrect based on the provided image.\\
            *****************************************\\
            Output Format:\\
            Answer: your answer\\
        \midrule
        \textbf{Prompts for hallucinating reasoning tasks:} \\Now, please help me generate new answers with hallucination errors based on the image, question, and answer provided. There are two cases now:\\
        1. If the given question and answer are short and do not require logical reasoning, then modify the answer to a hallucination error answer, such as some quantity errors or entity and property errors.\\
        2. If the entire question requires logical reasoning, then help me reorganize the answers based on the given image, questions, and answers into the format ``Reason: xxx, Result: xxx" (Answer 1). Modify the reasons by introducing errors related to logical relationships, entity information, entity attributes, etc. If the error in the reason would lead to a new result, modify the result accordingly. If the error does not lead to a new result, keep the original result. Similarly, organize it in the format ``Reason: xxx, Result: xxx" (Answer 2).\\
            *****************************************\\
            Question-answer Pair:\\
            Q: \{question\}\\
            A: \{answer\}\\
            
            Requirements:\\
            (1) The generated wrong answer and reasoning process should be combined with the image and be misleading..\\
            *****************************************\\
            Output Format:\\
            Answer: your answer\\
   \bottomrule
    \end{tabularx}
    \label{prompt}
\end{table}

%%%%%%%%%%%%%%%%%%%%%%%%%%%%%%%%%%%%%%%%%%%%%%%%%%%%%%%%%%%%%%%%%%%%%%%%%%%%%%%
%%%%%%%%%%%%%%%%%%%%%%%%%%%%%%%%%%%%%%%%%%%%%%%%%%%%%%%%%%%%%%%%%%%%%%%%%%%%%%%

\section{Details about baselines}
\label{sec:base}
For the assessment of multitasking, MME \cite{fu2023mme} is a comprehensive evaluation tool for MLLMs, which is designed to measure both perception and cognition abilities across 14 sub-tasks. MMBench \cite{liu2023mmbench} is characterized by its approach to assessing both perception and reasoning abilities, categorized into top-level dimensions in the ability taxonomy. This benchmark includes different levels of abilities, each including specific aspects of perception and reasoning. MM-Vet \cite{yu2023mm} focuses on six core capabilities for evaluation: recognition, knowledge, optical character recognition (OCR), spatial awareness, language generation, and math. These capabilities cover a wide range of functions from general visual recognition to specific tasks like arithmetic problem-solving. LLaVA-Bench \cite{liu2023visual} assesses models in more complex tasks and their adaptability to new domains. It consists of 24 diverse images, encompassing a variety of scenes like indoor and outdoor settings, memes, paintings, and sketches. Each image in LLaVA-Bench is paired with a detailed, manually crafted description and a carefully chosen set of questions, totaling 60 questions. This setup aims to provide a thorough and varied evaluation of the models' capabilities.

%% file: main.bbl
\begin{thebibliography}{45}
\providecommand{\natexlab}[1]{#1}
\providecommand{\url}[1]{\texttt{#1}}
\expandafter\ifx\csname urlstyle\endcsname\relax
  \providecommand{\doi}[1]{doi: #1}\else
  \providecommand{\doi}{doi: \begingroup \urlstyle{rm}\Url}\fi

\bibitem[Antol et~al.(2015)Antol, Agrawal, Lu, Mitchell, Batra, Zitnick, and Parikh]{antol2015vqa}
Antol, S., Agrawal, A., Lu, J., Mitchell, M., Batra, D., Zitnick, C.~L., and Parikh, D.
\newblock Vqa: Visual question answering.
\newblock In \emph{Proceedings of the IEEE international conference on computer vision}, pp.\  2425--2433, 2015.

\bibitem[Bai et~al.(2023)Bai, Bai, Yang, Wang, Tan, Wang, Lin, Zhou, and Zhou]{bai2023qwen}
Bai, J., Bai, S., Yang, S., Wang, S., Tan, S., Wang, P., Lin, J., Zhou, C., and Zhou, J.
\newblock Qwen-vl: A frontier large vision-language model with versatile abilities.
\newblock \emph{arXiv preprint arXiv:2308.12966}, 2023.

\bibitem[Bai et~al.(2022)Bai, Jones, Ndousse, Askell, Chen, DasSarma, Drain, Fort, Ganguli, Henighan, et~al.]{bai2022training}
Bai, Y., Jones, A., Ndousse, K., Askell, A., Chen, A., DasSarma, N., Drain, D., Fort, S., Ganguli, D., Henighan, T., et~al.
\newblock Training a helpful and harmless assistant with reinforcement learning from human feedback.
\newblock \emph{arXiv preprint arXiv:2204.05862}, 2022.

\bibitem[Bradley \& Terry(1952)Bradley and Terry]{bradley1952rank}
Bradley, R.~A. and Terry, M.~E.
\newblock Rank analysis of incomplete block designs: I. the method of paired comparisons.
\newblock \emph{Biometrika}, 39\penalty0 (3/4):\penalty0 324--345, 1952.

\bibitem[Cha et~al.(2023)Cha, Kang, Mun, and Roh]{cha2023honeybee}
Cha, J., Kang, W., Mun, J., and Roh, B.
\newblock Honeybee: Locality-enhanced projector for multimodal llm.
\newblock \emph{arXiv preprint arXiv:2312.06742}, 2023.

\bibitem[Chen et~al.(2023)Chen, Shen, Shao, Deng, and Nie]{chen2023lion}
Chen, G., Shen, L., Shao, R., Deng, X., and Nie, L.
\newblock Lion: Empowering multimodal large language model with dual-level visual knowledge.
\newblock \emph{arXiv preprint arXiv:2311.11860}, 2023.

\bibitem[Chuang et~al.(2023)Chuang, Jampani, Li, Torralba, and Jegelka]{chuang2023debiasing}
Chuang, C.-Y., Jampani, V., Li, Y., Torralba, A., and Jegelka, S.
\newblock Debiasing vision-language models via biased prompts.
\newblock \emph{arXiv preprint arXiv:2302.00070}, 2023.

\bibitem[Cui et~al.(2023)Cui, Zhou, Yang, Wu, Zhang, Zou, and Yao]{cui2023holistic}
Cui, C., Zhou, Y., Yang, X., Wu, S., Zhang, L., Zou, J., and Yao, H.
\newblock Holistic analysis of hallucination in gpt-4v (ision): Bias and interference challenges.
\newblock \emph{arXiv preprint arXiv:2311.03287}, 2023.

\bibitem[Dai et~al.(2023)Dai, Li, Li, Tiong, Zhao, Wang, Li, Fung, and Hoi]{instructblip}
Dai, W., Li, J., Li, D., Tiong, A. M.~H., Zhao, J., Wang, W., Li, B., Fung, P., and Hoi, S.
\newblock Instructblip: Towards general-purpose vision-language models with instruction tuning, 2023.

\bibitem[Dewangan et~al.(2023)Dewangan, Choudhary, Chandhok, Priyadarshan, Jain, Singh, Srivastava, Jatavallabhula, and Krishna]{dewangan2023talk2bev}
Dewangan, V., Choudhary, T., Chandhok, S., Priyadarshan, S., Jain, A., Singh, A.~K., Srivastava, S., Jatavallabhula, K.~M., and Krishna, K.~M.
\newblock Talk2bev: Language-enhanced bird's-eye view maps for autonomous driving.
\newblock \emph{arXiv preprint arXiv:2310.02251}, 2023.

\bibitem[Fu et~al.(2023)Fu, Chen, Shen, Qin, Zhang, Lin, Yang, Zheng, Li, Sun, et~al.]{fu2023mme}
Fu, C., Chen, P., Shen, Y., Qin, Y., Zhang, M., Lin, X., Yang, J., Zheng, X., Li, K., Sun, X., et~al.
\newblock Mme: A comprehensive evaluation benchmark for multimodal large language models.
\newblock \emph{arXiv preprint arXiv:2306.13394}, 2023.

\bibitem[Gao et~al.(2023)Gao, Schulman, and Hilton]{gao2022scaling}
Gao, L., Schulman, J., and Hilton, J.
\newblock Scaling laws for reward model overoptimization.
\newblock 2023.

\bibitem[Huang et~al.(2023)Huang, Dong, Zhang, Wang, He, Wang, Lin, Zhang, and Yu]{huang2023opera}
Huang, Q., Dong, X., Zhang, P., Wang, B., He, C., Wang, J., Lin, D., Zhang, W., and Yu, N.
\newblock Opera: Alleviating hallucination in multi-modal large language models via over-trust penalty and retrospection-allocation.
\newblock \emph{arXiv preprint arXiv:2311.17911}, 2023.

\bibitem[Jia et~al.(2021)Jia, Yang, Xia, Chen, Parekh, Pham, Le, Sung, Li, and Duerig]{jia2021scaling}
Jia, C., Yang, Y., Xia, Y., Chen, Y.-T., Parekh, Z., Pham, H., Le, Q., Sung, Y.-H., Li, Z., and Duerig, T.
\newblock Scaling up visual and vision-language representation learning with noisy text supervision.
\newblock In \emph{International conference on machine learning}, pp.\  4904--4916. PMLR, 2021.

\bibitem[Lee et~al.(2023)Lee, Phatale, Mansoor, Lu, Mesnard, Bishop, Carbune, and Rastogi]{lee2023rlaif}
Lee, H., Phatale, S., Mansoor, H., Lu, K., Mesnard, T., Bishop, C., Carbune, V., and Rastogi, A.
\newblock Rlaif: Scaling reinforcement learning from human feedback with ai feedback.
\newblock \emph{arXiv preprint arXiv:2309.00267}, 2023.

\bibitem[Leng et~al.(2023)Leng, Zhang, Chen, Li, Lu, Miao, and Bing]{leng2023mitigating}
Leng, S., Zhang, H., Chen, G., Li, X., Lu, S., Miao, C., and Bing, L.
\newblock Mitigating object hallucinations in large vision-language models through visual contrastive decoding.
\newblock \emph{arXiv preprint arXiv:2311.16922}, 2023.

\bibitem[Li et~al.(2023{\natexlab{a}})Li, Ge, Li, and Shan]{li2023visionlanguage}
Li, C., Ge, Y., Li, D., and Shan, Y.
\newblock Vision-language instruction tuning: A review and analysis.
\newblock 2023{\natexlab{a}}.

\bibitem[Li et~al.(2023{\natexlab{b}})Li, Wong, Zhang, Usuyama, Liu, Yang, Naumann, Poon, and Gao]{li2023llava}
Li, C., Wong, C., Zhang, S., Usuyama, N., Liu, H., Yang, J., Naumann, T., Poon, H., and Gao, J.
\newblock Llava-med: Training a large language-and-vision assistant for biomedicine in one day.
\newblock \emph{arXiv preprint arXiv:2306.00890}, 2023{\natexlab{b}}.

\bibitem[Li et~al.(2022)Li, Li, Xiong, and Hoi]{li2022blip}
Li, J., Li, D., Xiong, C., and Hoi, S.
\newblock Blip: Bootstrapping language-image pre-training for unified vision-language understanding and generation.
\newblock In \emph{International Conference on Machine Learning}, pp.\  12888--12900. PMLR, 2022.

\bibitem[Li et~al.(2023{\natexlab{c}})Li, Li, Savarese, and Hoi]{li2023blip}
Li, J., Li, D., Savarese, S., and Hoi, S.
\newblock Blip-2: Bootstrapping language-image pre-training with frozen image encoders and large language models.
\newblock \emph{arXiv preprint arXiv:2301.12597}, 2023{\natexlab{c}}.

\bibitem[Li et~al.(2023{\natexlab{d}})Li, Xie, Li, Chen, Wang, Chen, Yang, Wang, and Kong]{li2023silkie}
Li, L., Xie, Z., Li, M., Chen, S., Wang, P., Chen, L., Yang, Y., Wang, B., and Kong, L.
\newblock Silkie: Preference distillation for large visual language models.
\newblock \emph{arXiv preprint arXiv:2312.10665}, 2023{\natexlab{d}}.

\bibitem[Li et~al.(2023{\natexlab{e}})Li, Yin, Li, Chen, Wang, Ren, Li, Yang, Xu, Sun, et~al.]{li2023m}
Li, L., Yin, Y., Li, S., Chen, L., Wang, P., Ren, S., Li, M., Yang, Y., Xu, J., Sun, X., et~al.
\newblock M$^3$it: A large-scale dataset towards multi-modal multilingual instruction tuning.
\newblock \emph{arXiv preprint arXiv:2306.04387}, 2023{\natexlab{e}}.

\bibitem[Li et~al.(2023{\natexlab{f}})Li, Du, Zhou, Wang, Zhao, and Wen]{li2023evaluating}
Li, Y., Du, Y., Zhou, K., Wang, J., Zhao, W.~X., and Wen, J.-R.
\newblock Evaluating object hallucination in large vision-language models.
\newblock \emph{arXiv preprint arXiv:2305.10355}, 2023{\natexlab{f}}.

\bibitem[Liu et~al.(2023{\natexlab{a}})Liu, Lin, Li, Wang, Yacoob, and Wang]{liu2023aligning}
Liu, F., Lin, K., Li, L., Wang, J., Yacoob, Y., and Wang, L.
\newblock Aligning large multi-modal model with robust instruction tuning.
\newblock \emph{arXiv preprint arXiv:2306.14565}, 2023{\natexlab{a}}.

\bibitem[Liu et~al.(2023{\natexlab{b}})Liu, Li, Wu, and Lee]{liu2023visual}
Liu, H., Li, C., Wu, Q., and Lee, Y.~J.
\newblock Visual instruction tuning.
\newblock \emph{arXiv preprint arXiv:2304.08485}, 2023{\natexlab{b}}.

\bibitem[Liu et~al.(2023{\natexlab{c}})Liu, Duan, Zhang, Li, Zhang, Zhao, Yuan, Wang, He, Liu, et~al.]{liu2023mmbench}
Liu, Y., Duan, H., Zhang, Y., Li, B., Zhang, S., Zhao, W., Yuan, Y., Wang, J., He, C., Liu, Z., et~al.
\newblock Mmbench: Is your multi-modal model an all-around player?
\newblock \emph{arXiv preprint arXiv:2307.06281}, 2023{\natexlab{c}}.

\bibitem[Lu et~al.(2022)Lu, Mishra, Xia, Qiu, Chang, Zhu, Tafjord, Clark, and Kalyan]{lu2022learn}
Lu, P., Mishra, S., Xia, T., Qiu, L., Chang, K.-W., Zhu, S.-C., Tafjord, O., Clark, P., and Kalyan, A.
\newblock Learn to explain: Multimodal reasoning via thought chains for science question answering.
\newblock In \emph{The 36th Conference on Neural Information Processing Systems (NeurIPS)}, 2022.

\bibitem[OpenAI(2023)]{OpenAI_GPT4_2023}
OpenAI.
\newblock Gpt-4 technical report.
\newblock \emph{ArXiv}, abs/2303.08774, 2023.
\newblock URL \url{https://arxiv.org/abs/2303.08774}.

\bibitem[Rafailov et~al.(2023)Rafailov, Sharma, Mitchell, Manning, Ermon, and Finn]{rafailov2023direct}
Rafailov, R., Sharma, A., Mitchell, E., Manning, C.~D., Ermon, S., and Finn, C.
\newblock Direct preference optimization: Your language model is secretly a reward model.
\newblock In \emph{Thirty-seventh Conference on Neural Information Processing Systems}, 2023.
\newblock URL \url{https://arxiv.org/abs/2305.18290}.

\bibitem[Rohrbach et~al.(2018)Rohrbach, Hendricks, Burns, Darrell, and Saenko]{rohrbach2018object}
Rohrbach, A., Hendricks, L.~A., Burns, K., Darrell, T., and Saenko, K.
\newblock Object hallucination in image captioning.
\newblock \emph{arXiv preprint arXiv:1809.02156}, 2018.

\bibitem[Schulman et~al.(2017)Schulman, Wolski, Dhariwal, Radford, and Klimov]{schulman2017proximal}
Schulman, J., Wolski, F., Dhariwal, P., Radford, A., and Klimov, O.
\newblock Proximal policy optimization algorithms, 2017.

\bibitem[Sun et~al.(2023)Sun, Shen, Cao, Liu, Li, Shen, Gan, Gui, Wang, Yang, et~al.]{sun2023aligning}
Sun, Z., Shen, S., Cao, S., Liu, H., Li, C., Shen, Y., Gan, C., Gui, L.-Y., Wang, Y.-X., Yang, Y., et~al.
\newblock Aligning large multimodal models with factually augmented rlhf.
\newblock \emph{arXiv preprint arXiv:2309.14525}, 2023.

\bibitem[Taori et~al.(2023)Taori, Gulrajani, Zhang, Dubois, Li, Guestrin, Liang, and Hashimoto]{alpaca}
Taori, R., Gulrajani, I., Zhang, T., Dubois, Y., Li, X., Guestrin, C., Liang, P., and Hashimoto, T.~B.
\newblock Stanford alpaca: An instruction-following llama model.
\newblock \url{https://github.com/tatsu-lab/stanford_alpaca}, 2023.

\bibitem[Touvron et~al.(2023{\natexlab{a}})Touvron, Lavril, Izacard, Martinet, Lachaux, Lacroix, Rozi{\`e}re, Goyal, Hambro, Azhar, et~al.]{touvron2023llama}
Touvron, H., Lavril, T., Izacard, G., Martinet, X., Lachaux, M.-A., Lacroix, T., Rozi{\`e}re, B., Goyal, N., Hambro, E., Azhar, F., et~al.
\newblock Llama: Open and efficient foundation language models.
\newblock \emph{arXiv preprint arXiv:2302.13971}, 2023{\natexlab{a}}.

\bibitem[Touvron et~al.(2023{\natexlab{b}})Touvron, Martin, Stone, Albert, Almahairi, Babaei, Bashlykov, Batra, Bhargava, Bhosale, et~al.]{touvron2023llama2}
Touvron, H., Martin, L., Stone, K., Albert, P., Almahairi, A., Babaei, Y., Bashlykov, N., Batra, S., Bhargava, P., Bhosale, S., et~al.
\newblock Llama 2: Open foundation and fine-tuned chat models.
\newblock \emph{arXiv preprint arXiv:2307.09288}, 2023{\natexlab{b}}.

\bibitem[Tu et~al.(2023)Tu, Cui, Wang, Zhou, Zhao, Han, Zhou, Yao, and Xie]{tu2023many}
Tu, H., Cui, C., Wang, Z., Zhou, Y., Zhao, B., Han, J., Zhou, W., Yao, H., and Xie, C.
\newblock How many unicorns are in this image? a safety evaluation benchmark for vision llms.
\newblock \emph{arXiv preprint arXiv:2311.16101}, 2023.

\bibitem[Vinyals et~al.(2015)Vinyals, Toshev, Bengio, and Erhan]{vinyals2015show}
Vinyals, O., Toshev, A., Bengio, S., and Erhan, D.
\newblock Show and tell: A neural image caption generator.
\newblock In \emph{Proceedings of the IEEE conference on computer vision and pattern recognition}, pp.\  3156--3164, 2015.

\bibitem[Wei et~al.(2022)Wei, Wang, Schuurmans, Bosma, Xia, Chi, Le, Zhou, et~al.]{wei2022chain}
Wei, J., Wang, X., Schuurmans, D., Bosma, M., Xia, F., Chi, E., Le, Q.~V., Zhou, D., et~al.
\newblock Chain-of-thought prompting elicits reasoning in large language models.
\newblock \emph{Advances in Neural Information Processing Systems}, 35:\penalty0 24824--24837, 2022.

\bibitem[Ye et~al.(2023)Ye, Xu, Xu, Ye, Yan, Zhou, Wang, Hu, Shi, Shi, et~al.]{ye2023mplug}
Ye, Q., Xu, H., Xu, G., Ye, J., Yan, M., Zhou, Y., Wang, J., Hu, A., Shi, P., Shi, Y., et~al.
\newblock mplug-owl: Modularization empowers large language models with multimodality.
\newblock \emph{arXiv preprint arXiv:2304.14178}, 2023.

\bibitem[Yin et~al.(2023)Yin, Fu, Zhao, Xu, Wang, Sui, Shen, Li, Sun, and Chen]{yin2023woodpecker}
Yin, S., Fu, C., Zhao, S., Xu, T., Wang, H., Sui, D., Shen, Y., Li, K., Sun, X., and Chen, E.
\newblock Woodpecker: Hallucination correction for multimodal large language models.
\newblock \emph{arXiv preprint arXiv:2310.16045}, 2023.

\bibitem[Yu et~al.(2023{\natexlab{a}})Yu, Yao, Zhang, He, Han, Cui, Hu, Liu, Zheng, Sun, and Chua]{2023rlhf-v}
Yu, T., Yao, Y., Zhang, H., He, T., Han, Y., Cui, G., Hu, J., Liu, Z., Zheng, H.-T., Sun, M., and Chua, T.-S.
\newblock Rlhf-v: Towards trustworthy mllms via behavior alignment from fine-grained correctional human feedback.
\newblock \emph{arxiv}, 2023{\natexlab{a}}.

\bibitem[Yu et~al.(2023{\natexlab{b}})Yu, Yao, Zhang, He, Han, Cui, Hu, Liu, Zheng, Sun, et~al.]{yu2023rlhf}
Yu, T., Yao, Y., Zhang, H., He, T., Han, Y., Cui, G., Hu, J., Liu, Z., Zheng, H.-T., Sun, M., et~al.
\newblock Rlhf-v: Towards trustworthy mllms via behavior alignment from fine-grained correctional human feedback.
\newblock \emph{arXiv preprint arXiv:2312.00849}, 2023{\natexlab{b}}.

\bibitem[Yu et~al.(2023{\natexlab{c}})Yu, Yang, Li, Wang, Lin, Liu, Wang, and Wang]{yu2023mm}
Yu, W., Yang, Z., Li, L., Wang, J., Lin, K., Liu, Z., Wang, X., and Wang, L.
\newblock Mm-vet: Evaluating large multimodal models for integrated capabilities.
\newblock \emph{arXiv preprint arXiv:2308.02490}, 2023{\natexlab{c}}.

\bibitem[Zhao et~al.(2023)Zhao, Wang, Ouyang, Dong, Wang, and He]{zhao2023beyond}
Zhao, Z., Wang, B., Ouyang, L., Dong, X., Wang, J., and He, C.
\newblock Beyond hallucinations: Enhancing lvlms through hallucination-aware direct preference optimization.
\newblock \emph{arXiv preprint arXiv:2311.16839}, 2023.

\bibitem[Zhou et~al.(2023)Zhou, Cui, Yoon, Zhang, Deng, Finn, Bansal, and Yao]{zhou2023analyzing}
Zhou, Y., Cui, C., Yoon, J., Zhang, L., Deng, Z., Finn, C., Bansal, M., and Yao, H.
\newblock Analyzing and mitigating object hallucination in large vision-language models.
\newblock \emph{arXiv preprint arXiv:2310.00754}, 2023.

\end{thebibliography}
